\documentclass{article}

\usepackage{arxiv}

\usepackage[T1]{fontenc}    
\usepackage{hyperref}       
\usepackage{url}            
\usepackage{booktabs}       
\usepackage{amsfonts}       
\usepackage{nicefrac}       
\usepackage{microtype}      
\usepackage{lipsum}
\usepackage{graphicx}
\title{Multiform Fonts-to-Fonts Translation via Style and Content Disentangled Representations of Chinese Character }

\author{
  Fenxi Xiao \\
  Department of Electrical Engineering\\
  Jinan University\\
  Guangzhou, China 510632 \\
  \texttt{xfx0203@163.com} \\
  \And 
  Jie Zhang \\
  Department of Electrical Engineering\\
  Jinan University\\
  Guangzhou, China 510632 \\
  \texttt{714167281@qq.com} \\
  \And
  Bo Huang \\
  Department of Electrical Engineering\\
  Jinan University\\
  Guangzhou, China 510632 \\
  \texttt{abohuang@jnu.edu.cn} \\
   \And
 Xia Wu \\
  Department of Electrical Engineering\\
  Jinan University\\
  Guangzhou, China 510632 \\
  \texttt{wuxia\_liao@qq.com} \\  
}

\begin{document}
\maketitle

\begin{abstract}
This paper mainly discusses the generation of personalized fonts as 
the problem of image style transfer. 
The main purpose of this paper is to design a network framework 
that can extract and re-combine the content and style of the characters. 
These attempts can be used to synthesize the entire set of fonts 
with only a small amount of characters. 
The paper combines various depth networks such as Convolutional Neural Network, 
Multi-layer Perceptron and Residual Network to find the optimal model 
to extract the features of the fonts character.
The result shows that those characters we have generated is very close to real characters,
using Structural Similarity 
index and Peak Signal-to-Noise Ratio evaluation criterions.
\end{abstract}


\section{Introduction}
Language is a unique symbol of human civilization. Writing system is the development of the language.
Writing system usually has many different fonts.  Unlike the English font library 
that has only 26 alphabets, 
Chinese font library contains tens of thousands of characters. There are also thousands of commonly 
used characters.

The design of Chinese fonts is a very time-consuming and laborious task, 
with the calligrapher manually writing a large number of Chinese characters as a benchmark.
It is necessary to find a automation method to help the font designing 
for Chinese character. The best way is that 
by manually designing several characters, 
we can directly synthesize the remaining Chinese characters with the same style of character.

In the last few years, the development of deep learning make it possiable in automatic 
image style transfer. Several researcher have intended to generate Chinese fonts by using
different deep learning method\cite{Sun2017,Lyu2018,Deng,guo_creating_2018,
chang_rewrite2_nodate,chang_generating_nodate,azadi_multi-content_nodate}.
 Jiang et.al. using a U-net model realize end-to-end
Chinese character mapping, that automatically generate the whole GB2312 font library
 that consists of 6763 Chinese characters from a small number of characters written by the user.
\cite{Jiang2017}. Jiang develop an efficient and generalized deep framework W-Net, 
that is capable of learning and generating any arbitrary characters 
sharing the style similar to the given single font character\cite{amari_w-net_2017}.
Sun et. al. also propose a variational auto-encoder framework 
to generate Chinese characters\cite{Sun2017}.

This paper mainly discusses the generation of personalized fonts as 
the problem of image style transfer. 
The main purpose of this paper is to design a network framework 
that can extract and re-combine the content and style of the characters. 
The paper combines various depth networks such as Convolutional Neural Network, 
Multi-layer Perceptron and Residual Network to find the optimal model 
to extract the features of the fonts character.

These attempts can be used to synthesize the entire set of fonts 
with only a small amount of style characters. 
The result shows that those characters we have generated is very close to  real characters,
using Structural Similarity 
index and Peak Signal-to-Noise Ratio evaluation criterions.

\section{Method Description}
Due to the complicated structures of Chinese characters, It is not easy transfer 
deep learning methods widely used in image synthesis to this job.
The big problem is that the style and content features of Chinese characters are 
complexly entangled. The present deep learning methods, such as cross-domain 
disentanglement\cite{Gonzalez-Garcia2018}, Dientangled Representation\cite{Lee2018},
U-Net model\cite{esser_variational_nodate}, are not well work on Chinese characters synthesis.

In this paper, our propose a multiple mapping model to 
generater new Chines character. Our method is 
different from the current moethod in unpaired font dataset training.
Several researcher have intended to generate Chinese fonts by using
different deep learning method\cite{Sun2017,Lyu2018,Deng,guo_creating_2018,
chang_rewrite2_nodate,chang_generating_nodate,azadi_multi-content_nodate}.
These methods most require paired fonts to be trained. 
The paired fonts  have same content and different style.

\subsection{Multimodal mapping model}

\begin{figure}
  \begin{center}
  \includegraphics[scale=0.8]{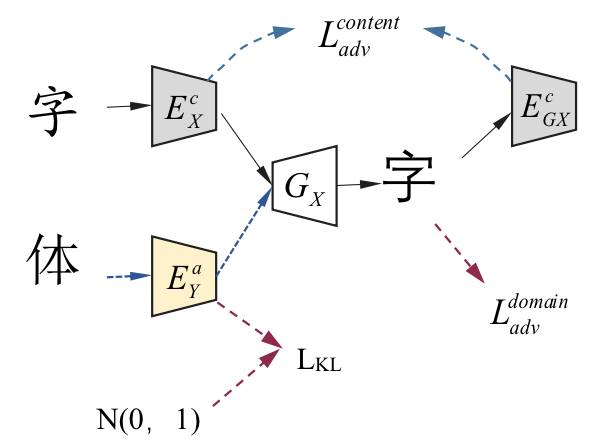}
  \caption{Multimodal mapping model}
  \label{fig1}
  \end{center}
\end{figure}

Our method is to learn a multimodal mapping between
two unpaired style fonts. Figure \ref{fig1} is a schematic of 
our multi-mapping model. 
Our model framework consists of a content encoder, 
a style encoder, a generator G, and a content discriminator.
The content encoder encodes the font image X into the content space, 
the style encoder encodes the font image Y 
into the style space.
Then stitches the content encoding of font X with the style encoding of font Y, 
and uses the generator G  to regenerate a new font image with the stiches vector.
The generator G obtains a latent feature representation 
from the encoder and outputs an image similar to the input images, which
mixed the content and style feature of the two input font. 
Then the generated font is encoded again by contend encoder to get
its content fetures.
The content discriminator distinguishes the extracted content 
between the input fonts and the generated fonts.

\begin{figure}
  \centering
  \includegraphics[scale=0.5]{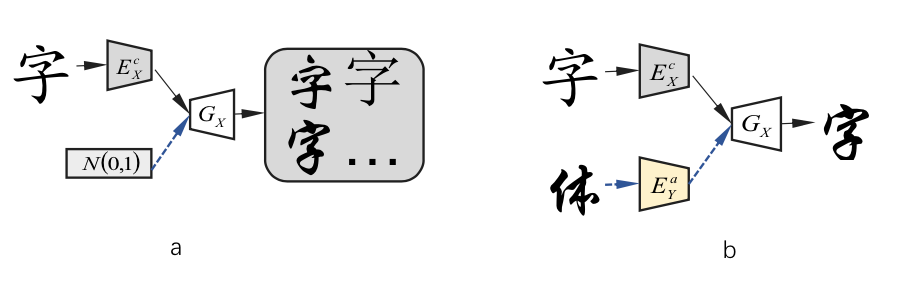}
  \caption{multiple types mapping model}
  \label{fig2}
\end{figure}

In order to obtain multiple types of fonts,
we normalize the style vector so that it can be sampled 
from the Gaussian prior probability N (0,1) or extracted from the input style fonts.
this process is shown in Figure \ref{fig2}. Figure \ref{fig2} (a) 
make randomly extracted style generate. and Figure \ref{fig2} (b) 
using a specified font style to generate new style fonts. 
Content discriminator forces the generator to produce a font with the same content.
the loss of the content discriminator is expressed as:

\begin{equation}
L_{adv}^{content}(E_X^c, E_{GX}^c, D^c) = \ \
E_x[\frac{1}{2} log D^c(E_X^c(x)) + \frac{1}{2} 
log(1-D^c(E_X^c(x)))] + 
E_x[\frac{1}{2}log D^c(E_{GX}^c(x)) + \frac{1}{2} log(1-D^c(E_{GX}^c(x)))]
\end{equation}

In addition, the adversarial loss and KL-Loss are added. 
the adversarial loss discriminator D tries to distinguish 
the real image and the generated image in each domain, 
while the generator G tries to generate a realistic image. 
The use of KL-Loss is to achieve random sampling during generate.
The KL-loss make the style representation as close 
as possible to the prior Gaussian distribution.
The all loss is expressed as :

\begin{eqnarray}
L_{all} = \lambda_{adv}^{content} L_{adv}^c + 
\lambda_{adv}^{domain} L_{adv}^{domain} + \lambda_{KL} L_{KL}
\end{eqnarray}

The different lambda represents the proportion of each loss function.
In the training, 
we set the parameters as follows:$\lambda_{adv}^{content}=1$,
$\lambda_{adv}^{domain}=1$, and $\lambda_{KL}=0.01$

\subsection{The implemented model}
\begin{figure}
  \centering
  \includegraphics[scale=0.5]{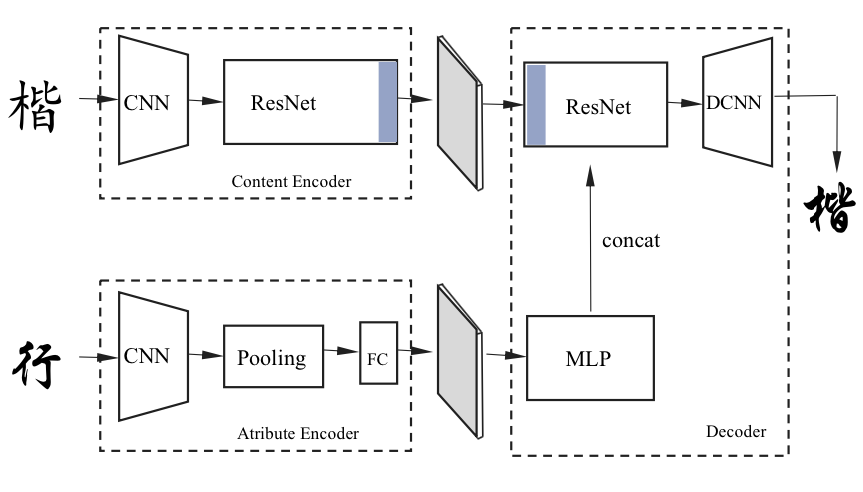}
  \caption{The detailed implemented model}
  \label{fig3}
\end{figure}

This model is implemented using the Tensorflow machine learning framework. 
The overall architecture consists of a content encoder, 
a style encoder, and a joint decoder.
It is difficult for a single encoder-decoder network to produce content and style features.
We use two encoders to extract content and style separately. 
Finally, we train the two subnets together. 
The overall model structure is shown in Figure \ref{fig3}.

The content encoder is composed of 3 convolutional layers 
and 4 residual modules. 
All of the convolutional layers are normalized.
 Weight sharing was introduced in the last layer of 
 the residual network and the first layer of the generator.

 The generator G is a decoder, 
 which reconstructs the input image according 
 to its content encoding and style encoding. 
 The decoder processes the content encoding 
 through a set of residual modules and 
 uses multiple up-sampled transposed convolutional layers 
 to generate a reconstructed image. 
 In addition, by using Affine Transformation 
 in the normalization layer to express style features, 
 We introduced the Adaptive Instance Normalization (AdaIN)\cite{Huang2017}.
The parameters in the AdaIN layer can be generated with
 the style coding by using multi-layer perceptron (MLP).

During the training phase, we use the Adam optimizer to randomly select batch-sized 
characters from the training set during each training batch processing step. 
The training period is set to 50, the number of iterations is 10,000, 
and the learning rate is 0.0001. The average training time is one week in a NVIDIA 2080 Ti Graphics.

\subsection{Train dataset}
Since the distribution of training data will have a 
great impact on the output of the convolutional neural network.
It is especially important to choose reasonable training data. 
Imbalanced training sets will have a negative impact on the results fonts.
 In order to ensure the balance of the training set, 
the train fonts dataset  is  compored of 3000 Chinese characters 
commonly used in the Chinese character library.
The content font in the training data set is a regular Kai style fonts, 
and the style font is 50 different styles. 
The whole train dataset contains about 300,000 fonts images. 

\section{Experiments results}
The following figure \ref{fig4} shows some training results. 
The left column is the content font, 
the middle column is the style font, and the right is the output result. 
It can be seen that the style and structure of the generater font is similar with the 
style fonts.

\begin{figure}
  \centering
  \includegraphics[scale=0.5]{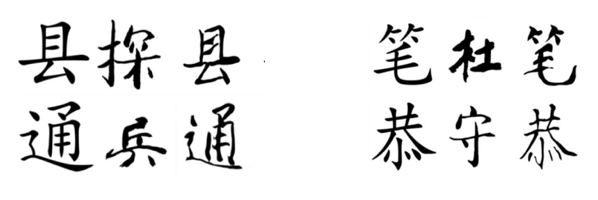}
  \caption{Unpaired training results}
  \label{fig4}
\end{figure}

\begin{figure}
  \centering
  \includegraphics[scale=0.5]{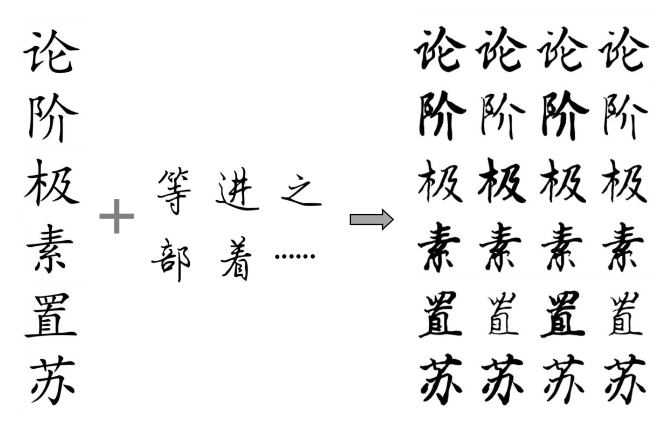}
  \caption{Converting images with desired fonts features}
  \label{fig5}
\end{figure}

In addition to randomly sampling the style information from the feature space, 
we can also use an image with the desired fonts style features for transformation.
Figure \ref{fig5} shown the generate fonts from desired fonts style.
The left column is the content font, 
the middle column is the desired style font, and the right is the generate fonts. 

\section{Conclusion}
In this paper,
we first propose a multi-mode mapping-based encoder-decoder 
framework that combines an automatic variational encoder 
and a generative adversarial network to learn unpaired font transform.
The model can extract a class of styles based on a given style fonts
 to generate an output image with one style. The model can 
 also   randomly extract multiple styles based on the style encoder
  to generate  multiple styles fonts.
   The fonts generated by the model are consistent with the real fonts.
 
\section*{Acknowledgement}
This work was supported by National Natural Science Foundation of China (61307080)
\bibliographystyle{unsrt}  

\bibliography{/Users/booq/Research/paper_by_me/zotero_bibtex}

\end{document}